\documentclass[journal]{IEEEtran}

\usepackage[table]{xcolor}
\usepackage{hhline}

\usepackage{bm,url,times,multirow,amsmath,amssymb,algorithm,xspace,epsfig,todonotes,array,color,todonotes,pifont,listings}
\usepackage{algorithm, mathtools}
\usepackage{caption}
\usepackage[noend]{algpseudocode}
\usepackage[bookmarks=false]{hyperref}

\usepackage{amssymb}
\usepackage{pifont}

\newcommand{\code}[1]{{\lstinline!#1!}}

\newcommand{\argmax}{\operatorname*{arg\,max}}

\newcolumntype{K}[1]{>{\centering\arraybackslash}p{#1}}

\newif\ifextrarefs
\extrarefsfalse

\definecolor{Gray}{gray}{0.9}
\definecolor{Gray2}{gray}{0.8}
\definecolor{Gray3}{gray}{0.6}

\begin{document}
%

\title{Rolling Horizon Evolutionary Algorithms for General Video Game Playing
\thanks{Raluca~D.~Gaina, Simon~M.~Lucas, Diego~P\'erez-Li\'ebana (School of Electronic Engineering and Computer Science, 10 Godward Square, Queen Mary University of London, Mile End Rd, London E1 4FZ, UK; email: {\tt \{r.d.gaina,simon.lucas,diego.perez\}@qmul.ac.uk}),
\newline
Sam Devlin (Microsoft Research Cambridge; email: {\tt Sam.Devlin@microsoft.com})}}

%
%
%

\author{Raluca~D.~Gaina, Sam Devlin, Simon~M.~Lucas, Diego~Perez-Liebana}

\markboth{IEEE Transactions on Games}%
{Shell \MakeLowercase{\textit{et al.}}: Bare Demo of IEEEtran.cls for Journals}

\maketitle

\begin{abstract} 

Game-playing Evolutionary Algorithms, specifically Rolling Horizon Evolutionary Algorithms, have recently managed to beat the state of the art in win rate across many video games. However, the best results in a game are highly dependent on the specific configuration of modifications and hybrids introduced over several papers, each adding additional parameters to the core algorithm. Further, the  best previously published parameters have been found from only a few human-picked combinations, as the possibility space has grown beyond exhaustive search. This paper presents the state of the art in Rolling Horizon Evolutionary Algorithms, combining all modifications described in literature, as well as new ones, for a large resultant hybrid. We then use a parameter optimiser, the N-Tuple Bandit Evolutionary Algorithm, to find the best combination of parameters in 20 games from the General Video Game AI Framework. 
Further, we analyse the algorithm's parameters and some interesting combinations revealed through the optimisation process. Lastly, we find  new  state  of  the  art  solutions  on  several  games by automatically exploring the large parameter space of RHEA.

\end{abstract}

\begin{IEEEkeywords}
Evolutionary computation, rolling horizon, computational intelligence, artificial intelligence, games, general video game playing, real-time games
\end{IEEEkeywords}

\IEEEpeerreviewmaketitle

\section{Introduction} \label{sec:intro}

In this paper we revisit the application of game-playing Evolutionary Algorithms with a deeper analysis of algorithm modifications and hybrids. We argue that automatic exploration (or algorithmic optimisation) of algorithm variations is essential for problems with large search spaces, although not exhaustive due to computation speed limitations. This optimisation process can further lead to insights into the algorithm being optimised, and as such we additionally conduct an in-depth analysis of the parameter space, while highlighting performance gain in various games.

There have been several recent advances in game-playing Evolutionary Algorithms~\cite{baier2018evolutionary,justesen2017playing} and a multitude of modifications and hybrids proposed to improve performance across a large number of games. The result is that the possibility space for algorithm configurations has grown beyond efficient manual optimisation. Although preforming grid-search is sometimes possible for finding good values for some parameters~\cite{gaina2017rhanalysis}, more recent works find the need to reduce more and more the number of parameter combinations chosen for analysis~\cite{gaina2019dynamic}. Therefore, the interesting insights into which variation of the algorithm is actually best are limited to human exploration of very small sections of the entirety of the search space. 

The specific novel application of Evolutionary Algorithms as game-playing methods (referred to as Rolling Horizon Evolutionary Algorithms, or RHEA) was introduced for the first time in 2013 by Perez et al.~\cite{Perez2013rheaptsp}. In the context of playing games, RHEA evolves, at every game step, a sequence of actions to play in the game; the first action of the best sequence found is played at the end of the evolutionary process and a new sequence is evolved for the subsequent game step. This base algorithm has been extended in several works. Gaina et al.~\cite{gaina2017rhanalysis} performed an in-depth analysis of the algorithm's main parameters (population size and individual length), generally finding that the higher the parameter values (even reaching the extreme of Random Search), the better RHEA performs across several games; this work further highlights an increase in performance with the increase of the available budget and correspondingly higher parameter values. Different population initialisation methods were explored in~\cite{gaina2017rhseeding}; this work was important in highlighting the benefit of using different options in different game types, as some games saw increased performance with greedy initialisation, while others preferred a statistical approach instead. Furthermore, Gaina et al. tested in~\cite{gaina2017rhhybrids} various hybrids and combinations with other techniques, which further pinpointed not only the difference in performance of certain parameter configurations across the different games, but also that the RHEA parameter space was already being expanded beyond the possibility of exhaustively exploring all parameter combinations. Some of these enhancements were further tested by Santos et al.~\cite{santos2018improved} in General Video Game AI (GVGAI) and by Tong et al.~\cite{tong2019enhancing} in MuJoCo's physical control tasks, both with great success. Finally, a study on dynamically adjusting individual length based on the fitness landscape observed during evolution~\cite{gaina2019dynamic} shows that some parameters might be conflicting with each other and cause poor performance in some games and suggests a need for carefully constructed parameter search spaces.

The work in this paper is carried out within the domain of general video game playing, which focuses on finding general-purpose Artificial Intelligence players that are able to play any game, even those unseen previously. The concepts behind this could be further extended to general AI which is able to solve any given task (as opposed to any given game), as methods developed for games have been shown to be applicable to wider domains, such as chemistry~\cite{segler2018planning}. Two large categories of players can be differentiated in this domain: planning and learning. The latter requires training for several episodes on a game before it can figure out how to play it, which is often an expensive process leading to narrow results: the agent trained for one game would be unlikely to be able to play another without significant training on the new game. The former category, which RHEA belongs to, refers to methods which work online, during the game, to search for the appropriate solutions. These methods require an internal model of the games (referred to as a forward model, or FM) to be able to simulate possible futures and effects of their actions. Although planning methods are more generally applicable, they face the drawback of the lack of a FM in some games, as this is not always feasible. The problem of learning any game's model is an active research area~\cite{schrittwieser2019mastering} which would make our methods even more widely applicable, even in complex commercial games; however, in this work we apply our algorithms to games which do have a model available.

In this context, Monte Carlo Tree Search (MCTS) had for a long time represented the state of the art in general video game playing. However, RHEA has been shown to outperform MCTS in multiple games in some of its variations~\cite{gaina2017rhhybrids}, while other combinations of modifications led to significantly worse results. As highlighted by Lucas et al.~\cite{lucas2019efficient}, there can be a large difference in performance for the same base algorithm when using different parameters, and optimisation is essential. Ashlock et al.~\cite{ashlock2017general} emphasise this in the context of general game playing, where one single method (or single parameter configuration, in our approach) is unlikely to achieve high performance across all possible tasks. Our specific problem is additionally highly noisy: most games are stochastic and the same sequence of actions in a game could lead to different outcomes; furthermore, the algorithm itself is stochastic and may produce different outputs given the same game state. 

In this paper, we use the N-Tuple Bandit Evolutionary Algorithm (NTBEA)~\cite{kunanusont2017n} for optimising RHEA parameters, an algorithm which has shown robust high performance in noisy optimisation problems, even when compared with alternatives. It features high sample efficiency, fast convergence and good scaling for large search spaces~\cite{lucas2019efficient}. Simulations of AI players on a multitude of games can be very expensive, therefore sample efficiency is key, making NTBEA suitable for optimising RHEA parameters. The algorithm has been previously successfully employed in several noisy optimisation problems, such as tuning game parameters~\cite{kunanusont2018scoretrend} as well as AI game-player parameters~\cite{lucas2018n,lucas2019efficient,bravi2019rinascimento}. A highly adaptive system which can optimise its parameters and structure so as to achieve best performance in various games could easily feed into a generic life-long learning system such as that presented in~\cite{gaina2019thyia}.


A summary of our contributions is as follows:
\begin{enumerate}
    \item We give an overview of the current state of the art in Rolling Horizon Evolutionary Algorithms within the context of general video game playing.
    \item We perform an in-depth analysis and optimisation of the algorithm's parameters with respect to its performance across the various games tested.
    \item We find new configurations which outperform the previous state of the art on a range of GVGAI games.
\end{enumerate}


\section{Background}\label{sec:bg}

This section describes the two key concepts employed in this paper, the Rolling Horizon Evolutionary Algorithm (RHEA) and the N-Tuple Bandit Evolutionary Algorithm (NTBEA), as well as introducing the framework and game set used for experiments.

\subsection{Rolling Horizon Evolution} \label{sec:rhea-bg}

RHEA utilises Evolutionary Algorithms (EA) to evolve an in-game sequence of actions at every game tick, with restricted computation time per execution. This subsection will describe the baseline algorithm, often referred to as \textit{vanilla}; modifications applied are detailed in Section~\ref{sec:methods}. 

In this application of EAs for game-playing, the genotype is described as a vector of integers of length $L$ (individual length), where each integer $a$ is in the range $[0, N)$, with $N$ being the maximum number of actions in a given game. This translates to a phenotype as a sequence of actions played in the game starting from state $S_0$, or, in other words, the behaviour of the player. If the action chosen at any game step is illegal or cannot reasonably be played (e.g. walking into a wall), it is automatically treated as \textit{``do nothing''} by the game engine, thus the EA does not see any individuals, or genes in the individuals, as illegal or infeasible. In order to evaluate an individual in this context, RHEA uses the forward model (FM) of the game, an internal model of the world, to simulate through the actions, one at a time. The game state reached at the end is then evaluated with a heuristic function $h$ and this value becomes the fitness of the individual: therefore, we are evolving action sequences which lead to the best game outcome, limited to the exploration range $L$. The heuristic function $h$ is always kept to a generic form throughout the experiments; this aims to maximize the game score, while favouring wins and discouraging losses, see Equation~\ref{eq:heuristic} ($s$ is the game state being evaluated and $score$ is the game score normalised in $(0,1)$).

\begin{equation}\label{eq:heuristic}
h(s) = \begin{cases}
1 & \text{win = True}\\
0 & \text{win = False}\\
score & \text{otherwise}
\end{cases}
\end{equation}

Using this method to evaluate individuals, the vanilla algorithm follows a typical EA process. It begins by initialising a population of $P$ individuals of length $L$ at random and evaluates them. At every generation, while budget is still available, it promotes $E$ individuals directly to the next generation through elitism. It then generates $P$ offspring by repeatedly selecting parents through tournament selection, crosses them with uniform crossover to create a child, and mutates the child through uniform mutation before adding it to the pool of offspring. The best $P-E$ individuals from both parents and offspring pools are added to the next generation and the process repeats. Typically, a budget of $40$ms per game tick is given to the algorithm for real-time decision-making. 

We would further like to highlight the benefits of choosing RHEA over Monte Carlo Tree Search (MCTS), as the current favourite in general game-playing methods: RHEA is easily parallelizable, with each individual being independent from the rest and from the search process. It is able to handle continuous actions without modification, which has been previously explored in~\cite{samothrakis2014rolling}. It can easily adapt to n-player games with minimal changes through co-evolution~\cite{Liu2016coevo} and the many possible enhancements and modifications make it highly adaptive and controllable for different given problems.



\begin{table*}[t]
\centering
\caption{Game set including feature analysis. The last 3 columns show clusters as depicted in previous works; games with the same value are denoted as part of the same cluster. As \cite{bontrager2016matching} do not include all of these games in their study, column \cite{nelson2016investigating} shows the game indexes between which the missing games are placed by Mark Nelson (lower-higher); \cite{stephenson2018continuous} shows more recent work clustering all GVGAI games.}
\begin{tabular}{|c|c||c|c|c|c|c|c|c|c||c|c|c|}
\hline
\textbf{Idx} & \textbf{Game}      & \textbf{Stoch.} & \textbf{Rewards} & \textbf{Win} & \textbf{Lose}   & \textbf{Levels} & \textbf{NPCs} & \textbf{Res.} & \textbf{Actions} & \textbf{\cite{bontrager2016matching}} & \textbf{\cite{nelson2016investigating}} & \textbf{\cite{stephenson2018continuous}} \\ \hline\hline
0 & Dig Dug & x & D & Puzzle/Kill & Timeout & L/Dense & E &  & Move+Shoot & 4 &  & 5 \\ \hline
1 & Lemmings &  & D & Exit/Puzzle & Death & L/Dense & N &  & Move+Shoot & 4 &  & 5 \\ \hline
2 & Roguelike & x & D & Exit & Death & L/Dense & E & x & Move+Shoot & 4 &  & 4 \\ \hline
3 & Chopper & x & D+Disq & Kill & No-kill & L/Dense & E & x & Move+Shoot &  & g4-g1 & 2 \\ \hline
4 & Crossfire & x & N & Exit & Death & M/Dense & E &  & Move & 2 &  & 4 \\ \hline
5 & Chase &  & D & Kill & Death & M/Sparse & F+E &  & Move & 2 &  & 4 \\ \hline
6 & Camel Race &  & N & Exit & Timeout & L/Sparse & E &  & Move & 2 &  & 3 \\ \hline
7 & Escape &  & N & Exit/Puzzle & Death & M/Dense &  &  & Move & 2 &  & 3 \\ \hline
8 & Hungry Birds &  & Disq & Exit & Timeout & M/Sparse &  & HP & Move &  & g7-g10 & 3 \\ \hline
9 & Bait &  & N & Puzzle/Exit & Timeout & S/Sparse &  & x & Move & 4 &  & 4 \\ \hline
10 & Wait for Breakfast &  & N & Puzzle & Timeout & M/Dense & N &  & Move & 2 &  & 3 \\ \hline
11 & Survive Zombies & x & D & Timeout & Death & M/Dense & F+E &  & Move & 3 &  & 4 \\ \hline
12 & Modality &  & N & Puzzle & Timeout & S/Dense &  &  & Move & 3 &  & 4 \\ \hline
13 & Missile Command &  & D+Disq & Kill & No-kill & M/Sparse & E &  & Move+Shoot & 3 &  & 2 \\ \hline
14 & Plaque Attack &  & D & Kill & No-kill & L/Dense & E &  & Move+Shoot & 3 &  & 2 \\ \hline
15 & Sea Quest & x & D+Disq & Timeout & Death & M/Dense & F+E & x & Move+Shoot & 3 &  & 2 \\ \hline
16 & Infection & x & D & Kill & Timeout & M/Dense & F+E &  & Move+Shoot & 1 &  & 1 \\ \hline
17 & Aliens & x & D & Kill & Death & M/Dense & E &  & LR+Shoot & 1 &  & 1 \\ \hline
18 & Butterflies & x & D & Kill & Timeout & M/Dense & F &  & Move & 1 &  & 2 \\ \hline
19 & Intersection & x & D+Disq & Timeout & Death & L/Dense & E & HP & Move &  & g18-g17 & 1 \\ \hline
\end{tabular}
\label{tab:games}
\end{table*}


\subsection{N-Tuple Bandit Evolutionary Algorithm}\label{sec:ntbea}


NTBEA is a model-based optimiser based on an Evolutionary Algorithm. It begins by randomly initiating a solution $o$, or with a given solution (referred to as \textit{seed}, and the process as \textit{seeding} the algorithm) if specified in its application. It then evaluates one solution at a time, with the evaluation method determined by the specific application, and adds it to its internal $n$-tuple model - that is, all combinations of $n$ parameters are registered to have observed the fitness of the evaluated solution. We use $1$, $2$ and $L$ tuples, where $L$ is the solution length.

$n = 50$ neighbours of the solution are then generated through uniform random mutation, with probability $1/L$, forming neighbourhood $N$. The fitness of all neighbours is estimated based on previously observed $n$-tuple values, by calculating: the average values of the neighbour's tuples, as well as the average number of times each tuple was previously explored. These two statistics are used within a bandit equation (see Equation ~\ref{eq:ucb-ntbea}) to choose the neighbour with the highest value to be the solution evaluated next. This equation aims to balance between promising solutions by exploiting high fitness values $Q(o)$ and uncertain solutions by exploring those seen less during the process $exp(o)$. The constant $k = 2$ sets the focus of the algorithm, whether more exploitative or more exploratory. Small random noise (maximum $\epsilon=0.5$) is added to each neighbour's final value to randomly break ties. 

The process then repeats with the new chosen solution ($o'$) for a set number of iterations.

\begin{equation}\label{eq:ucb-ntbea}
o' = \argmax_{o \in N} \left\{Q(o) + k \times exp(o) + noise\right\}
\end{equation}

\subsection{Framework} \label{sec:framework}

We use NTBEA to tune RHEA parameters within the General Video Game AI (GVGAI) framework~\cite{gvgaibook2019}. GVGAI is a framework widely used in research~\cite{perez2018gvgaisurvey} which features a corpus of over $100$ single-player games and $60$ two-player games. These are fairly small games, each focusing on specific mechanics or skills the players should be able to demonstrate, including clones of classic arcade games such as Space Invaders, puzzle games like Sokoban, adventure games like Zelda or game-theory problems such as the Iterative Prisoners Dilemma. All games are real-time and require players to make decisions in only $40$ms at every game tick, although not all games explicitly reward or require fast reactions; in fact, some of the best game-playing approaches add up the time in the beginning of the game to run Breadth-First Search in puzzle games in order to find an accurate solution~\cite{perez2018gvgaisurvey}. However, given the large variety of games (many of which are stochastic and difficult to predict accurately), scoring systems and termination conditions, all unknown to the players, highly-adaptive general methods are needed to tackle the diverse challenges proposed.

GVGAI includes several different tracks which tackle different problems: single-player planning~\cite{perez2016general}, two-player planning~\cite{gainaGVGAI2Pb} and single-player learning tracks focus on finding general game-playing AI agents which would be capable of planning (with internal models of the world) or learning across all the games in the framework. More recently, level generation~\cite{khalifa2016general} (creating levels for any game) and rule generation~\cite{khalifa2017general} (creating rules for any given level) challenges were introduced as well, to push the limits of general game AI.

For the purpose of the experiments described in this paper, we will focus on the single-player planning track, although the work could easily be expanded to include two-player games.

\subsection{Game set} \label{subsec:games}

We select $20$ single-player games out of the larger GVGAI corpus, as previously analysed in several works. First introduced in~\cite{gaina2017rhanalysis} and described in Table~\ref{tab:games}, the game set used in this study is sampled based on GVGAI competition entries performance across large subsets, so as to include games of varying difficulty. Additionally, half of the games are deterministic and half are stochastic, introducing additional noise to the parameter optimisation problem explored in this paper.

The game table includes additional information about each game. They showcase varying reward structures, such as games with \textit{no} rewards (with the possibility of gaining points on win/lose conditions only), games with \textit{dense} rewards (multiple interactions with the environment result in a score change) or games with \textit{discontinuous} rewards (a longer sequence of actions is required to obtain the reward). Four different types of winning conditions are featured, in which the player has to kill certain game objects (\textit{Kill}), reach an exit point (\textit{Exit}), wait for a timer to run out (\textit{Timeout}) or complete a certain more precise sequence of actions (\textit{Puzzle}, such as move a box onto a specific point). Three types of losing conditions are included, which result in the player losing if they run out of time (\textit{Timeout}), die (\textit{Death}) or fail to kill specific game objects (\textit{No-kill}).

Additionally, the 5 levels included with each game vary in \textit{size} (Large - \textit{L}, Medium - \textit{M} or Small - \textit{S}) and \textit{density} of interactive tiles (that is, tiles which produce some sort of effect when the player interacts with it, such as blocking the player's path, moving or getting destroyed). Some games include Non-Player Character (NPCs) that might either help the player (\textit{F}), hurt the player (\textit{E}) or have no direct influence on the player's win/lose condition or score (\textit{N}) through their behaviour. The player may need to collect resources or pay particular attention to their avatar's hitpoints (HP). Finally, games vary in the actions available to the players (\textit{Move} includes movement in all 4 directions, up, down, left and right; \textit{LR} includes only left and right movement; a special \textit{Shoot} action might be available in some games, with different effects). All symbols mentioned here refer strictly to the table notation.

When discussing parameter choices, we will refer to games as similar based on the features described in Table~\ref{tab:games}, or the clustering identified from previous works 


\section{RHEA Parameter Space}\label{sec:methods}

\begin{figure}[t]
\centering
\includegraphics[width=0.99\columnwidth]{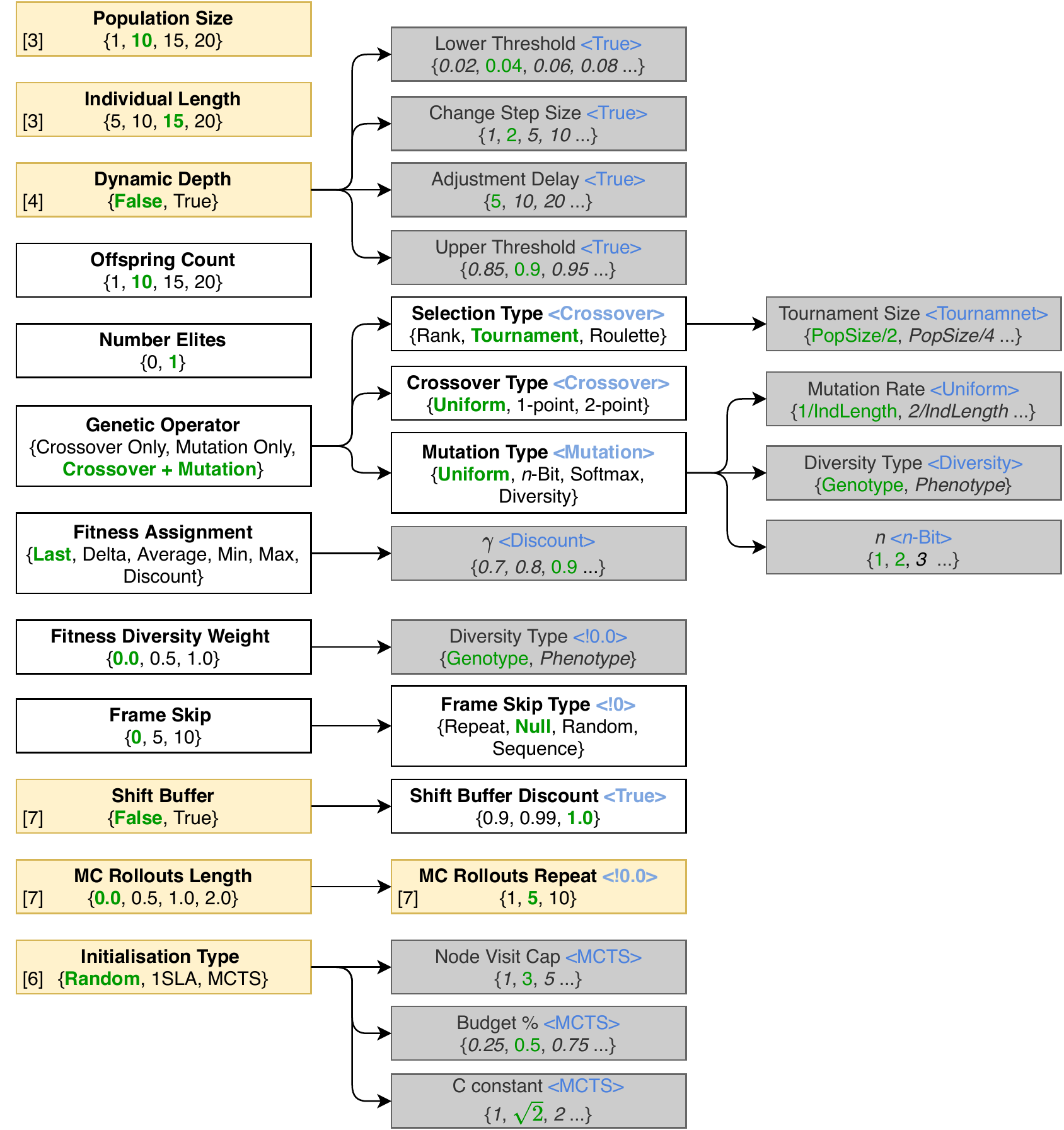}
\caption{Parameter search space, total size $4.03 \times 10^8$. Possible values for each parameter are between curly brackets, with default value highlighted in bold green. Parameters not in the $1^{st}$ column are dependent on others (denoted with arrows from parent to dependent; parent value required for dependent to affect phenotype is noted in blue, between angled brackets). Greyed-out parameters are not included in the experiments for this paper (the default value were used instead). In yellow parameters previously analysed in literature, with citation.}
\label{fig:params}
\end{figure}

This section describes all the evolutionary algorithm hyper-parameters used for the experiments, including hybrids and game-specific modifications, some introduced in previous work~\cite{gaina2017rhanalysis,gaina2017rhseeding,gaina2017rhhybrids,gaina2019dynamic}, as highlighted in Figure~\ref{fig:params}. Dependent parameters (2nd and 3rd column) are parameters that would not impact the phenotype without specific values taken by parent parameters, as detailed below.

\subsection{Genetic operators}

There are three main genetic operators used by the evolutionary algorithm in RHEA: crossover, selection and mutation. In our implementation, selection is only used to select parents for offspring, subsequent generations being formed directly with the best individuals from the current generation (with no further selection being applied). These three genetic operators each have several implementation options, as discussed below. A hyper-parameter controls which operators should be applied, with options of only using crossover (and selection), only using mutation, or using all three to first obtain an offspring from crossover and then mutate it as well. It is worth noting that the operator type parameters are dependent on the choice of genetic operator: changing the mutation type parameter would not have any effect on the phenotype if no mutation is used in the algorithm, and similarly for crossover and selection.

\textbf{Selection.} Three types of selection are available in the system: tournament, roulette and rank. Tournament selection picks a percentage of the population ($t=50\%$) randomly and then chooses the best individuals from these to reproduce. Roulette selection chooses individuals with probabilities equal to their fitness (therefore, higher fitness individuals have a higher chance of being selected). Rank selection first assigns inverse-ranks to all individuals in the population according to their fitness (the lowest fitness individual would have rank 1, second lowest rank 2 etc.) and then choose individuals with probabilities equal to their rank (therefore, higher fitness individuals have a higher chance of being selected, but the selection pressure is reduced by minimizing the differences in fitness).

\textbf{Crossover.} Two types of crossover are available in the system: uniform and $n$-point. Uniform crossover selects genes from either of the parents with equal probability. $n$-point crossover randomly selects $n$ points along the individuals which would split all individuals in subsections, the offspring being formed then by alternatively choosing subsections of genes from the parents; we use $1$ and $2$ as possible values for $n$, leading to three total values for the crossover parameter.

\textbf{Mutation.} Four types of mutation are available in the system: uniform, softmax, diversity and $n$-bits. Uniform mutation assigns each gene an equal probability of mutation ($m=1/L$, where $L$ is the individual length) and picks a different value for the genes mutating uniformly at random. Softmax mutation uses the softmax equation (see Equation~\ref{eq:softmax}) to bias mutation towards the beginning of the individual, which causes the largest perturbation in the action sequence (changing any gene in the individual, in this context, also changes the meaning of all subsequent genes - therefore changes in the beginning of the genome have the largest impact in the phenotype). Diversity mutation keeps track of all values for all genes from all individuals explored during evolution and chooses to mutate the gene that has currently been explored the least, to the value for the gene that has been explored the least. Finally, $n$-bit mutation chooses $n$ genes uniformly at random to mutate to a new and different random value; we use $1$ and $2$ as possible values for $n$, leading to a total of five values for the mutation parameter.

\begin{equation}\label{eq:softmax}
    \text{Softmax}(x_{i}) = \frac{\exp(x_i)}{\sum_j \exp(x_j)}
\end{equation}

\subsection{Fitness assignment}

A key part of evaluating individuals, represented as action sequences, is the fitness assignment resulting from the phenotype interpretation (i.e. a sequence of game states the AI player traverses through the action sequence). If all game states traversed are evaluated with a heuristic function $h$ (see Equation~\ref{eq:heuristic}), then this array of values $F$ corresponding to each of the game states can be translated to a fitness value in different ways: keeping only the value of the last game state reached: $F[L-1]$; keeping the difference between the value of the last state and the value of the first state, so state improvement value: $\Delta(F[L-1],F[0])$; keeping the average of all game state values: $\overline{F}$; keeping the minimum value, a pessimistic model: $min(F)$; keeping the maximum value, an optimistic model: $max(F)$; or keeping a discounted sum of all values: $\sum_{i=0}^{L}F[i]\times\gamma^i$, where $\gamma=0.9$, which prioritises immediate rewards.

\subsection{Initialisation}

In the vanilla version, the algorithm is initialised with random individuals (all genes in all individuals are picked uniformly at random from all possible values). Different initialisation (or seeding) methods have been previously tested in conjunction with the vanilla algorithm with various success~\cite{gaina2017rhseeding}. Both One Step Look Ahead (1SLA) and Monte Carlo Tree Search (MCTS) initialisation options, which have shown promise in various games in the previous study by Gaina et. al are included in this system.

\textbf{1SLA.} This algorithm performs an exhaustive search of all possible actions in a given game state and picks the action which leads to the highest value for the following game state (evaluated with heuristic $h$ and breaking ties randomly). To form an individual in the EA, this process is followed for each gene: the best action is chosen for the first gene, and the game state is advanced with the chosen action; this process is then repeated again for all subsequent genes until an action sequence of sufficient length is generated. If the end of the game is reached during the creation of an individual, the individual is padded with random actions until it meets the required length. For initialisation of a RHEA population, the first individual is created with the 1SLA algorithm and the rest become mutated from the first. Given the greedy approach, this reduces (and often completely removes) the randomness of the initial population, in order to begin search from a local optimum.

\textbf{MCTS.} This algorithm iteratively builds a search tree by selecting nodes in the tree to expand using the UCB1 formula, see Equation~\ref{eq:ucb1}, where: constant $C=\sqrt[]{2}$, $a$ is the chosen action from the set of possible actions $A(s)$, $s$ is the current game state, $Q(s,a)$ is the value of choosing action $a$ from state $s$, $N(s)$ is the number of times state $s$ has been visited and $N(s,a)$ is the number of times state $s$ has been visited and action $a$ was chosen next. It then evaluates nodes with Monte Carlo simulations (a sequence of random actions up to a maximum tree depth $L$) starting from the newly expanded node and updates the statistics ($N(s)$, $N(s,a)$ and $Q(s,a)$ of all nodes traversed during an iteration with the value given by the heuristic $h$ for the final game state reached after Monte Carlo simulations. This tree grows asymmetrically as MCTS balances between exploration of uncertain actions and exploitation of seemingly good actions. For initialisation of a RHEA population, MCTS is run for half of the entire thinking budget of the agent (leaving half available for evolution), and the first individual is selected by greedily traversing the tree created. As the tree would not be fully expanded, the path through the tree is capped when a node with less than $3$ visits is reached and actions are added randomly up until individual length $L$; the rest of the individuals are mutated from the first.

\begin{equation}\label{eq:ucb1}
a^* = \argmax_{a \in A(s)} \left\{Q(s,a) + C \sqrt{\frac{ \ln N(s) }{ N(s,a) }}\right\}
\end{equation}

\subsection{Frame skip}

Frame skipping has become common practice in several Reinforcenemnt Learning works, and key in the success of specific applications~\cite{braylan2015frame,machado2018revisiting}: grouping $N$ game states when making a decision, to increase the data available and reduce the frequency of decisions returned to only every $N$ game states. Statistical forward planning approaches, on the other hand, usually make a new decision at every game tick, repeating their search process in the very limited time. With this modification, we test if SFP methods can also benefit from a longer time for making decision by only returning an action every $N$ game ticks, replying according to a specific strategy for the game ticks inbetween and using all the time inbetween decisions for planning the next move. We test $0$ (no frame skip, decisions at every game tick), $5$ and $10$ as values for $N$ and four different strategies for actions inbetween decisions: \textit{repeat}, \textit{null}, \textit{random} and \textit{sequence}. The \textit{repeat} strategy simply repeats the previously decided action until a new action is decided. The \textit{null} strategy plays \texttt{ACTION\_NIL} (does nothing), which more closely mimics human player gameplay with pauses inbetween actions. The \textit{random} strategy plays a random action and the \textit{sequence} strategy continues playing the following actions in the best individual returned with the last decision. The frame skip type parameter is dependant on the frame skip value: if no frame skip is used (value $0$), then changing the frame skip type would have no effect on the phenotype.

A form of frame skip using the repeat strategy described above was previously tested in GVGAI by Perez et al.~\cite{perez2017macro-physics} with notable success in several games.

\subsection{Shift buffer}

This is a population management technique which avoids repeating the entire search process from scratch at every new game tick, which usually loses information gained in previous iterations of the algorithm; this is meant to make the algorithm more sample-efficient by retaining previous computation information. The shift buffer has been employed in several works and tested in GVGAI by Gaina et al.~\cite{gaina2017rhhybrids}, and it works by keeping the final population evolved during one game tick to the next. However, as the first action of the best individual has just been played, all first actions from all individuals in the population are removed and a new random action is added at the end (a possible extension to this would keep only the individuals with the same first action as the one chosen). Additionally, there exists the option in our implementation to apply a discount to the values of all individuals in the new population, which can be either $0.9$, $0.99$ or $1.0$ (no discount applied); this would weaken the values of previously obtained sequences in the new context. The shift buffer discount parameter is dependent on the shift buffer toggle: if no shift buffer is used, then changing the shift buffer discount would have no effect on the phenotype.

\subsection{Dynamic depth}

It is often the case that different games benefit from different algorithm parameters. In particular, the individual length has a high impact in the performance of the vanilla RHEA, as shown in~\cite{gaina2017rhanalysis}. This was mainly tied to the density of rewards in the various games in~\cite{gaina2019dynamic}: games with dense rewards generally benefit from shorter individuals which would allow for more generations and more statistics gathered to facilitate quick strategic reactions; as opposed to games with sparse or no rewards, where longer individuals are required in order to be able to find those rewards further ahead. This difference in reward density can also be observed at a more granular level, during the play-through of only one game: some areas of the game may contain more rewards, whereas others would require more exploration. Therefore, the dynamic depth modification presented in~\cite{gaina2019dynamic} is included in the system, which has the option to change the length of the individuals at every $5$ game ticks: if the standard deviation of the fitness landscape observed previously falls below a threshold ($0.04$), decisions are considered to be uncertain without much variety in rewards observed and the individual length is increased by $2$; if the opposite happens and the fitness landscape observed previously raises above a threshold ($0.9$), more generations are prioritised for more informed decision making in a highly varied environment and the individual length is decreased by $2$ instead. All parameters for dynamic length adjustments were set based on~\cite{gaina2019dynamic} and could represent one point for further increasing the parameter search space in further studies.

\subsection{MC rollouts}

Lastly, we consider the hybridisation of the algorithm and its further combination with MCTS, which has been very successful in many GVGAI games~\cite{nelson2016investigating}. We have previously described MCTS initialisation, but concepts from MCTS can further be borrowed and integrated into RHEA, such as its Monte Carlo (MC) simulation phase. As described in~\cite{gaina2017rhhybrids}, the evaluation process in RHEA may add MC rollouts of length \{$0.0$ (no rollouts used), $0.5$, $1.0$ or $2.0$\} $\times$ $L$ after advancing through the action sequence of length $L$ represented by the individual; these may be repeated $1$, $5$ or $10$ times for more statistics gathered. In order for this to be compatible with the fitness assignment modifications, the values of all game states traversed (or the average value for a particular game tick if there are repetitions performed) $R$ are added at the end of the array of state values $F$ obtained from the individual and all fitness assignment methods are applied to the combined array of values ($F + R$) instead. The MC rollout repetition parameter is dependent on the rollout length: if the length is set to $0.0$, then changing the number of rollout repetitions would have no effect on the phenotype.


\section{Experiments}\label{sec:exp}

Given the large number of parameter combinations, estimated at $4.03 \times 10^8$, it would take a significant amount of time to test each combination exhaustively in several games and with repetitions for statistical significance. Therefore we choose to analyse the different parameters indirectly through the evolutionary process described by an N-Tuple Bandit Evolutionary Algorithm (NTBEA). We ran NTBEA for $1500$ iterations on each of the $20$ games described in Section~\ref{sec:bg} to perform a search through the RHEA parameter space depicted in Figure~\ref{fig:params}. Each individual evaluated by NTBEA would therefore be one parameter combination ($18$ individual length). We seed NTBEA with the previous state-of-the-art parameter configuration for each game (see `SotA' rows in Table~\ref{tab:perf}). To evaluate each individual, we run RHEA with the specific parameter configuration on the given game, once in each of the $5$ levels of the game and we use the average win rate on the $5$ levels as individual fitness. To test the final configuration, we run it $100$ times on the given game ($20$ times per level) and we additionally test the tuned parameter configuration on the entire set of $20$ games, similarly with $100$ runs per game.

All experiments were run on IBM System X iDataPlex dx360 M3 Server nodes, with one game per node, having one Intel Xeon E5645 processor core allocated to it and a maximum of 3GB of RAM of JVM Heap Memory. The runs took between $43$ hours and $6$ days to complete, including NTBEA tuning and final configuration testing; one run of a game can take up to $2000$ game ticks to complete, with $1000$ Forward Model calls per tick for AI decision making (plus game engine computations), the fastest game ending after $50$ game ticks on average. The budget for all agents was set as $1000$ Forward Model calls instead of time limits (which averages as the equivalent of $40$ms in our tests), in order for the experiments to be consistent and replicable across different machines.

In the following sections we aim to analyse not only the performance of the optimised agents on the different games, but also the parameter space explored during the evolution and the parameter choices themselves. We hypothesise that similar games would lead to similar choices in parameters, which would differ across game types.

The paper presents and discusses the most interesting aspects observed, but all results, plotting scripts and additional figures are available on Github\footnote{\url{https://github.com/rdgain/ExperimentData/tree/NTBEA-RHEA-2019}}.


\begin{figure}[t]
\centering
\includegraphics[width=0.49\columnwidth]{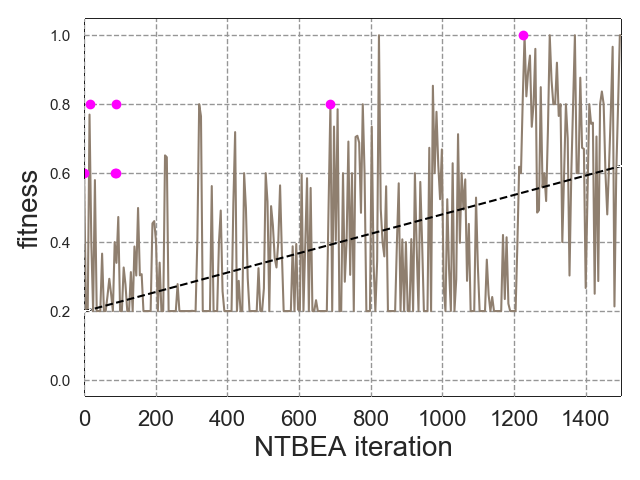}
\includegraphics[width=0.49\columnwidth]{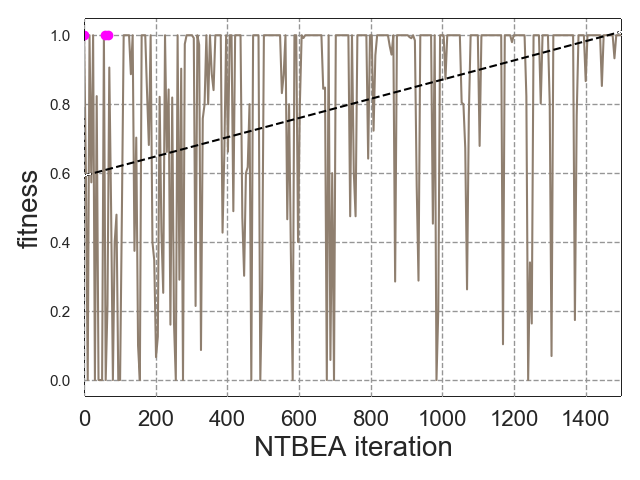}
\caption{Progression of solution fitness during NTBEA optimisation process in 2 games, \textit{Missile Command} (left) and \textit{Intersection} (right). The pink dots indicate when the solution evaluated becomes the new best (the last pink dot before or at iteration X is the solution which NTBEA would return if execution was stopped at iteration X). Trendline in black.}
\label{fig:fit}
\end{figure}

\subsection{Optimisation Effectiveness}

We first discuss the effectiveness of the optimisation. We summarise in Table~\ref{tab:perf} the results obtained on all 20 games used for tuning RHEA parameters with NTBEA. For each game, we present the parameter configuration of the previous state of the art (previous highest win rate recorded), its win rate and standard error; similarly, we present the optimised configuration for each game.

There are many games in which the win rate remains at or very close to $0\%$. This set of games (\textit{Dig Dug}, \textit{Lemmings}, \textit{Roguelike}) remains too difficult for these methods to solve without more game-specific information or better exploration policies. 

There are also several games which see win rates at, or very close to, $100\%$ (\textit{Intersection}, \textit{Aliens}, \textit{Infection}, \textit{Chopper} and \textit{Plaque Attack}). We do not see a decrease in performance in these games after optimisation (but a definite increase in \textit{Plaque Attack} to $100\%$ with several modifications in parameter choices, including using dynamic depth, 1SLA initialisation and random frame skip).

We do see several games improving performance significantly: the win rate in \textit{Sea Quest} increases from $65\%$ to $84\%$ by employing longer individual lengths, a larger population size, a shift buffer and MC rollouts. \textit{Missile Command} sees an increase in win rate from $77.78\%$ to $86\%$ with a shift buffer, MC rollouts and a discounted fitness assignment. And performance in \textit{Camel Race} increases from $11\%$ to $41\%$ by using \textit{repetition} frame skip, a shift buffer and MC rollouts, amongst many configuration modifications. These 3 games do not immediately show common features as per Table~\ref{tab:games}, with \textit{Sea Quest} standing out due to its stochastic nature and dense environment, while the other two feature sparser deterministic environments.

We see a decrease in performance in three of the games: \textit{Butterflies} (from $96\%$ to $90\%$), \textit{Escape} (from $46\%$ to $32\%$) and \textit{Modality} (from $37.5\%$ to $25\%$). As NTBEA was seeded with the previously best solution, we believe these are cases in which the noisy fitness evaluation was shown to be most harmful, as the initial solution ended up with a worse fitness than the solutions returned. However, with more runs of the two configurations on the game, their rank turns out to be opposite. A similar smaller decrease is also observed in \textit{Lemmings} and \textit{Bait} - all of these, except for \textit{Butterflies}, are games with puzzle elements to them, which appear to be most difficult to optimize and estimate solution quality for, as they require more precise action sequences, with one move possibly making the game unsolvable, and therefore more precise evaluation.

Finally, we highlight NTBEA's optimisation process progression in two games in Figure~\ref{fig:fit}, \textit{Missile Command} and \textit{Intersection}. Both of these games see an upwards trend in solution quality, and they represent the games with the slowest and fastest convergence, respectively. We can observe that the algorithm settles on the solution for \textit{Intersection} very quickly, before iteration $100$, whereas it uses almost all computation budget for \textit{Missile Command} to find the best option. This could be an indication of not only game difficulty, but also strategic depth: most parameter options work well and obtain very good performance in \textit{Intersection}, while \textit{Missile Command} poses a challenge at which not many options are successful and the finding of those few good configurations is more difficult.

Overall, the game-specific optimised agents achieve win-rates of below $50\%$ when tested on the entire set of games, which is not surprising in the general game playing context; the agents do not use any game-specific information. The best performing tuned agent is that for \textit{Sea Quest} ($53.4$ average win rate on all $20$ games), which shares different features with several other games; this appears to make the specific configuration more generally applicable than the others.


\begin{figure}[t]
\centering
\includegraphics[height=0.3\columnwidth]{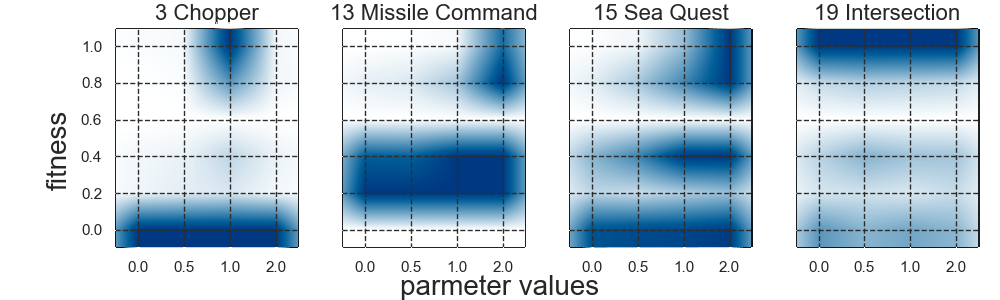}
\caption{1-tuple: rollout length percentage parameter. Color intensity represents number of occurrences of the data point (the darker the color, the more occurrences).}
\label{fig:rollouts}
\end{figure}

\subsection{1-tuple Analysis}

We further look into the parameter space explored by NTBEA starting with 1-tuples: that is, looking at each parameter in isolation and its preferred values in the different games tested; we use the term \textit{prefer} to mean the value achieves highest win rate. We group together the solutions in which the parameter had the same value chosen and plot the fitness of these solutions against parameter values. It is worth noting that it possible the parameter may have not had a great influence in the fitness obtained. We further exclude the data points where the parameter had no influence at all, in the case of dependent parameters (see Figure~\ref{fig:params}). The resulting heatmaps show the fitness values observed for each parameter values, as well as how many times each parameter value was explored by NTBEA. The latter is given by the intensity of colours in the figures presented; we cap the maximum number of occurrences of a data point at $100$ and normalize all values in [$0, 1$] for visualisation purposes.

$17$ games prefer the shift buffer turned on and to keep 1 elite between generations. Additionally, as previously seen in~\cite{gaina2017rhanalysis}, most games prefer long individual length and large population sizes. $17$ games further prefer the agent employing Monte Carlo rollouts at the end of its individual evaluation: \textit{Chopper}, \textit{Sea Quest} and \textit{Missile Command} in particular prefer very long rollouts ($2.0 \times L$, see Figure~\ref{fig:rollouts}); this could be due to these three games featuring different types of rewards and delays in obtaining rewards. The other similar game in terms of rewards, \textit{Intersection}, does not show a particular preference in this parameter, achieving $1.0$ fitness in all values. Full plots and results are available on GitHub.

\begin{figure}[t]
\centering
\includegraphics[height=0.3\columnwidth]{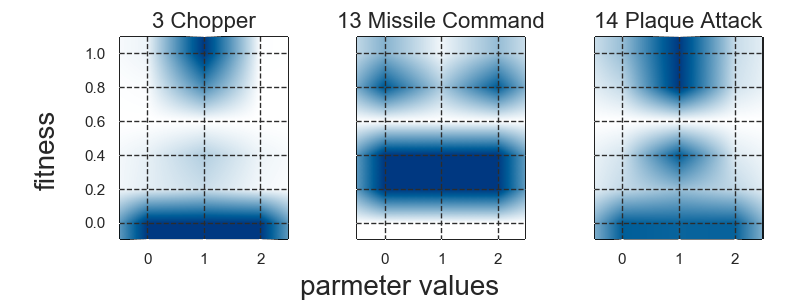}
\caption{1-tuple: genetic operator parameter. 0 - crossover and mutation. 1 - mutation only. 2 - crossover only.}
\label{fig:genop}
\end{figure}

In terms of genetic operators, most games prefer the agent to use both mutation and crossover in its evolutionary process. However, there are some exceptions: \textit{Chopper} and \textit{Plaque Attack} prefer to use mutation only, whereas \textit{Missile Command} prefers options that do include crossover and more disturbance in its offspring (see Figure~\ref{fig:genop}). Although these games are seen as similar in~\cite{stephenson2018continuous} and obtain high winning rates, the way the agents achieve their good performance does differ in these games, suggesting win-rate-based clustering methods could be improved by taking into account agent-based features.

\begin{figure}[t]
\centering
\includegraphics[height=0.6\columnwidth]{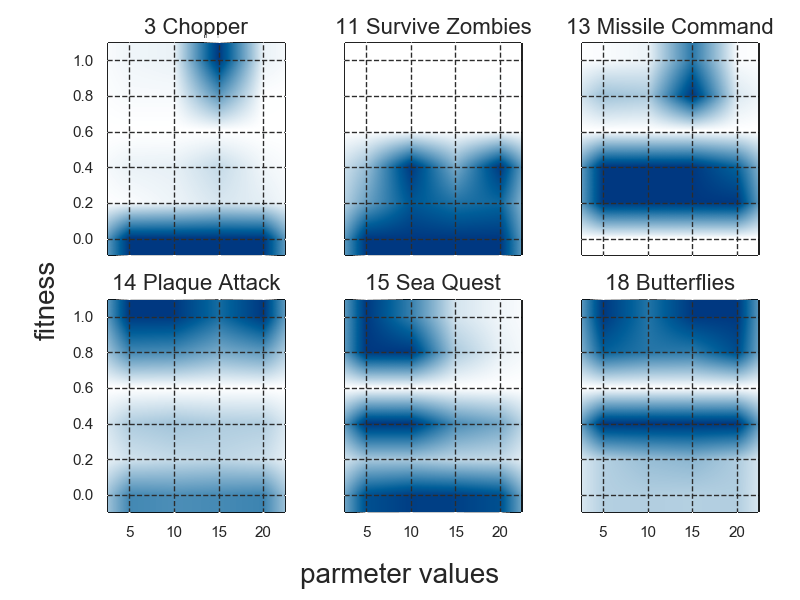}
\caption{1-tuple: offspring count parameter.}
\label{fig:offspring}
\end{figure}

When looking at the number of offspring (see Figure~\ref{fig:offspring}), \textit{Survive Zombies}, \textit{Missile Command} and \textit{Chopper} prefer more. These games are quite similar in terms of features (win/lose conditions, level sizes, enemy NPCs) and are clustered together in~\cite{stephenson2018continuous}. However, in the same cluster, \textit{Butterflies} and \textit{Plaque Attack} don't show strong preference here - as opposed to the others, these two games have a smoother reward function and score progression, while \textit{Missile Command} and \textit{Chopper} show more delay in getting rewards, more actions are required from the player to find particular rewarding scenarios. \textit{Sea Quest} is placed in a similar cluster by~\cite{bontrager2016matching}, but it shows opposite preference, for less offspring instead. In this game we see large discontinuous rewards as well as many smaller dense rewards - the larger variety in types of rewards could be what leads to favouring less solutions sampled to increase the number of generations in the evolutionary process, and to gain better insight into which reward type is preferable.

\begin{figure}[t]
\centering
\includegraphics[height=0.3\columnwidth]{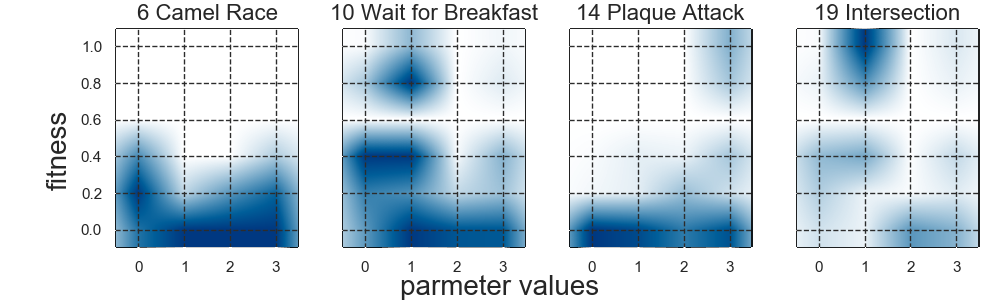}
\caption{1-tuple: frame skip type parameter. 0 - repeat. 1 - null. 2 - random. 3 - sequence.}
\label{fig:frameskip}
\end{figure}

Lastly, we highlight that \textit{Intersection} and \textit{Wait for Breakfast} are the only games that benefit largely from \textit{null} frame skipping (see Figure~\ref{fig:frameskip}) - in both of these games it is essential to wait for specific events to happen (a way in the road to clear, or the waiter to arrive) and this is highlighted in choice of method. \textit{Plaque Attack} prefers \textit{sequence} frame skipping, as plans evolved are precise enough in line with the constant stream of rewards. And \textit{Camel Race} prefers \textit{repeat} or \textit{sequence} frame skipping, with more frames skipped being better, which are more effective strategies of exploring large sparse environments. Most other games dislike frame skipping and prefer more fine-grained search; however, we note that the choice in values for this parameter is very coarse and it might be that more games could benefit from \textit{some} or dynamic frame skipping.

\begin{figure}[t]
	\centering
\includegraphics[width=0.95\columnwidth]{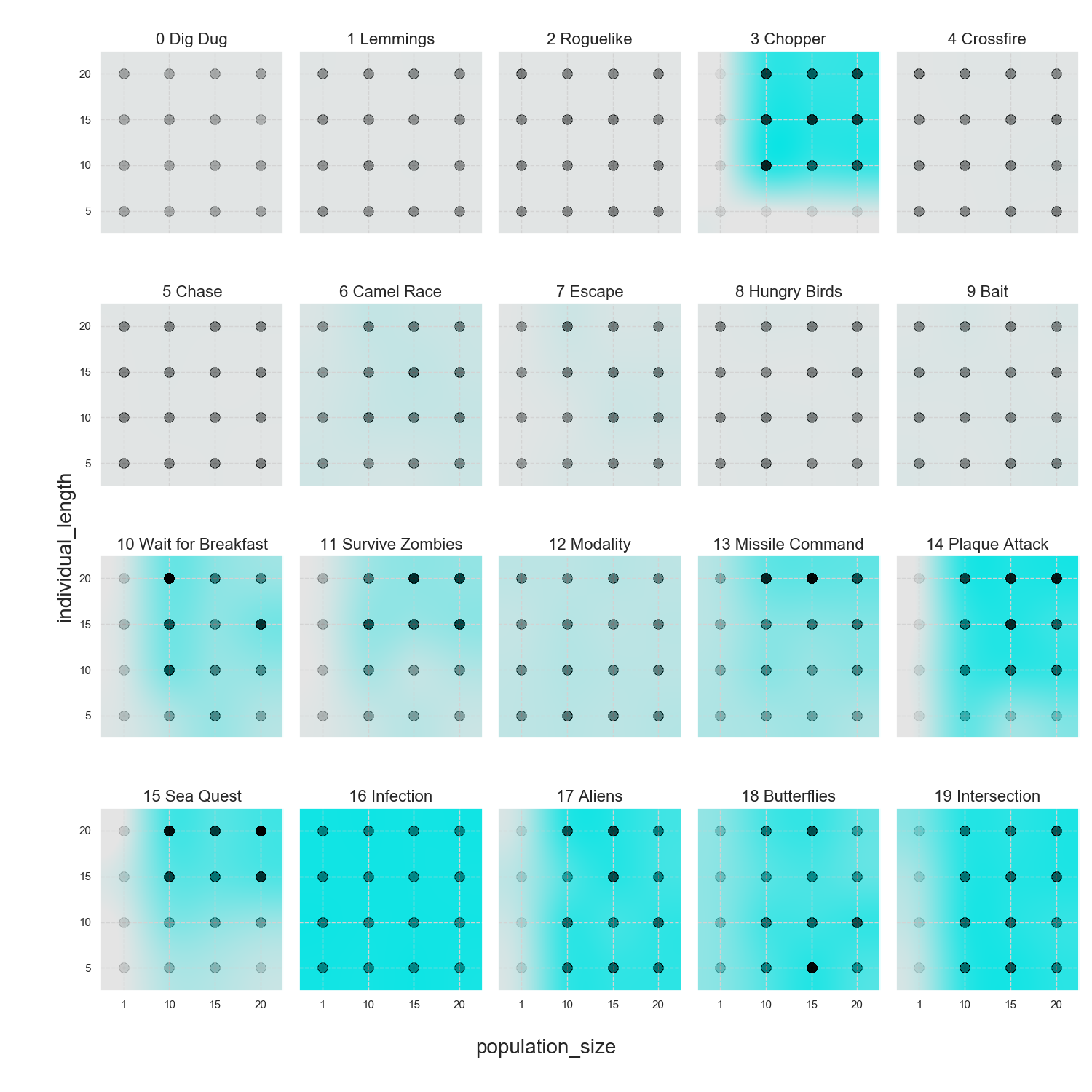}
    \caption{2-tuple: individual length and population size. Colors show average fitness for each data point, with blue being highest ($1.0$) and gray being lowest ($0.0$). Each data point is highlighted with a black circle; the darker the circle, the more number of times that combination of values was sampled.}
    \label{fig:ind-pop}
\end{figure}

\subsection{2-tuples Analysis}

Similarly as with 1-tuples, we can look at how combinations of parameters affect overall solution fitness. In this section we group together solutions which had the same values for each parameter combination, while eliminating the data points where either one or both of the parameter values did not impact solution phenotype, in case of dependent parameters (see Figure~\ref{fig:params}). We plot each parameter against all others in the different games tested, each data point representing the average fitness observed in the respective group of solutions. We further add black circles on each data point to highlight the number of times each combination was explored by NTBEA during the optimisation process.

The first thing that stands out in all resulting figures is that NTBEA explores the best combinations the most, while mostly ignoring less promising options. This can be seen as a direct confirmation of the effectiveness of the bandit-based approach, but also as a potential point of improvement: due to the nature of the very noisy optimisation, it might be beneficial to obtain more accurate estimates of some data points which do not immediately stand out as the best: as discussed in a previous section, it was the case in several games that the optimised solution ended up performing worse than the initial solution given to the algorithm, which could have been avoided had a more accurate evaluation of solution quality been done.

In Figure~\ref{fig:ind-pop} we can observe the combination of individual length $L$ and population size $P$ parameters. We've previously observed that longer individuals lead to higher fitness values, and similar for larger population sizes. It is interesting to see that this holds true also for the combination of $L$ and $P$, although specific combinations achieve better results in some games (such as $L=20$ and $P=15$ in \textit{Chopper}).

\begin{figure}[t]
	\centering
\includegraphics[width=0.85\columnwidth]{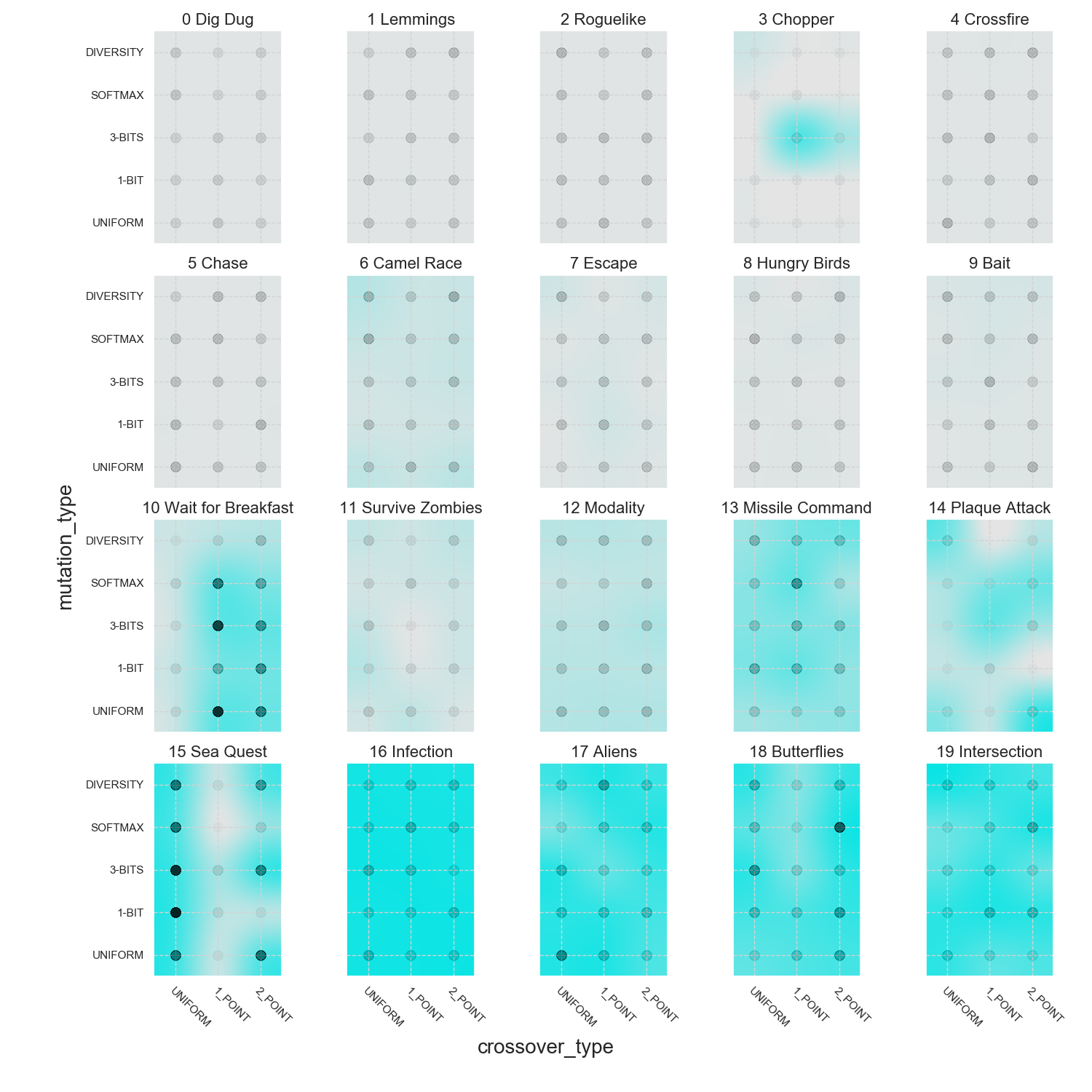}
    \caption{2-tuple: mutation type and crossover type.}
    \label{fig:mut-cross}
\end{figure}

Another interesting parameter combination to discuss is that of mutation type and crossover type, shown in Figure~\ref{fig:mut-cross}, which largely decides how offspring are created at each generation. Although the overall fitness of solutions differs, games \textit{Hungry Birds} and \textit{Plaque Attack} show a similar distribution of good or bad quality combinations: in particular, 1-point crossover does not agree with diversity mutation, and 2-point crossover does not agree with bit-mutation. This could largely be due to the specific modifications n-point crossover wishes to generate, which are modified unexpectedly by bit-mutation. However, these two games singled out here do not appear to have much in common according to our feature descriptions and clustering in Table~\ref{tab:games}; it is thus interesting to find game similarities beyond those given by traditionally-employed features.

\section{Conclusion}\label{sec:con}

In this paper we use the N-Tuple Bandit Evolutionary Algorithm (NTBEA) to optimise the performance of Rolling Horizon Evolutionary Algorithm (RHEA) in 20 GVGAI games, by modifying the configuration of RHEA's 18 parameters. The various values possible for all parameters form a large search space of $4.03 \times 10^8$, which makes manual optimisation or exhaustive search difficult with limited compute, thus we choose to use NTBEA to attempt to improve the win rate of the agent in each of the $20$ games.

As a result of the optimisation, the performance increases in several games. However, puzzles appeared to be the games where NTBEA struggled to estimate the quality of different agent configurations and the solution returned was worse than the state of the art, although NTBEA's evolutionary process was run with SotA as the initial solution. The optimisation process differed in the games tested, NTBEA being able to converge in under $100$ iterations in some games, while taking most of its $1500$ iteration budget to find good solutions in others: this strengthens the idea that one specific method is unlikely to perform well across all games, and that games might require specialised parameter search spaces to ensure fast optimisation or even the possibility of a high-performing solution being found.

We further analysed RHEA's parameters through the evolutionary process, by looking at some 1-tuples and 2-tuples and the values explored for each. Several games with similar features in common were found to prefer similar parameter values, although exceptions do exist of game clusters shown in parameter values, but not in the traditional game features considered. This suggests that game clustering methods can be further enhanced by considering agent-based features.

To further expand on the work carried out in this paper, we propose further exploring larger and more complex search spaces, with an enhanced NTBEA which is able to handle tree-structures: we've seen several parameters dependent on others and optimisation would be more sample-efficient if this was taken into account during the evolutionary process. More enhancements can also be added into the system, as well as optimising RHEA on a larger set of games (including multi-player games and games with hidden information), with the possibility of testing approaches at optimising a generally applicable player. Lastly, information gathered during optimisation and in-depth analysis can be used for designing hyper-parameter methods which would be able to identify game features, relate these to previously seen situations and adapt to new unknown environments.

\section*{Acknowledgments}
This work was funded by the EPSRC CDT in Intelligent Games  and Game Intelligence (IGGI)  EP/L015846/1

\begin{table*}[!t]
\centering
\caption{RHEA best win rate (and standard error) recorded in all games. ``opt'' rows show NTBEA optimisation results, other rows show previously best recorded (with corresponding citation, highlighted in yellow). Parameters using default values (as per Figure~\ref{fig:params}) highlighted in green. Enhancements include values for dependants in brackets. Win-rates in bold are the higher values observed, if different.}
\setlength\tabcolsep{2pt}
\setlength\extrarowheight{2pt}
\begin{tabular}{|c|c|cccccccccc|c}
\cline{1-12}
\multirow{3}{*}{\cellcolor{gray!10}} & \multirow{3}{*}{\cellcolor{gray!10}} & \multicolumn{10}{c|}{\cellcolor{gray!10}\textbf{Parameters}} &  \\ \cline{3-12}
\cellcolor{gray!10} & \cellcolor{gray!10} & \multicolumn{4}{c|}{\cellcolor{gray!10}\textbf{Numerical}} & \multicolumn{5}{c|}{\cellcolor{gray!10}\textbf{Nominal}} & \cellcolor{gray!10} &  \\ \cline{3-11}
\multirow{-3}{*}{\cellcolor{gray!10}\textbf{Game}} & \multirow{-3}{*}{\cellcolor{gray!10} \textbf{Win Rate}} & \cellcolor{gray!10}\textbf{P.Size} & \cellcolor{gray!10}\textbf{I.Len} & \cellcolor{gray!10}\textbf{Offspring} & \multicolumn{1}{c|}{\cellcolor{gray!10}\textbf{Elite}} & \cellcolor{gray!10}\textbf{Init.} & \cellcolor{gray!10}\textbf{Selection} & \cellcolor{gray!10}\textbf{Crossover} & \cellcolor{gray!10}\textbf{Mutation} & \cellcolor{gray!10}\textbf{Fit.} & \multicolumn{1}{|c|}{\multirow{-2}{*}{\cellcolor{gray!10}\textbf{Enhancements}}} &  \\ \hline

\cellcolor{gray!10} & 0\% $(0.00)$ & \cellcolor{green!25}10 & \cellcolor{green!25}15 & \cellcolor{green!25}10 & \multicolumn{1}{c|}{\cellcolor{green!25}1} & \cellcolor{green!25}RND & \cellcolor{green!25}Tourn. & \cellcolor{green!25}Uniform & \cellcolor{green!25}Uniform & \cellcolor{green!25}Last & \multicolumn{1}{|c|}{\cellcolor{green!25}-} & \multicolumn{1}{c|}{\cellcolor{yellow!25}\cite{gaina2017rhanalysis}} \\ \cline{2-13} 
\multirow{-2}{*}{\cellcolor{gray!10}0} & 0\% $(0.00)$ & 1 & 20 & 15 & \multicolumn{1}{c|}{\cellcolor{green!25}1} & MCTS & Rank & \cellcolor{green!25}Uniform & 2-bit & \cellcolor{green!25}Last & \multicolumn{1}{|c|}{S.Buff(0.9); MC(0.5,1); F.Skip(Rep)} & \multicolumn{1}{c|}{opt} \\ \hline

\cellcolor{gray!10} & \textbf{4\% $(1.98)$} & \cellcolor{green!25}10 & \cellcolor{green!25}15 & \cellcolor{green!25}10 & \multicolumn{1}{c|}{\cellcolor{green!25}1} & \cellcolor{green!25}RND & \cellcolor{green!25}Tourn. & \cellcolor{green!25}Uniform & \cellcolor{green!25}Uniform & \cellcolor{green!25}Last & \multicolumn{1}{|c|}{DD} & \multicolumn{1}{c|}{\cellcolor{yellow!25}\cite{gaina2019dynamic}} \\ \cline{2-13} 
\multirow{-2}{*}{\cellcolor{gray!10}1} & 0\% $(0.00)$ & \cellcolor{green!25}10 & 10 & 1 & \multicolumn{1}{c|}{0} & \cellcolor{green!25}RND & \cellcolor{green!25}Tourn. & 1-point & 2-bit & \cellcolor{green!25}Last & \multicolumn{1}{|c|}{\begin{tabular}[c]{@{}c@{}}S.Buff(0.9); MC(0.5,5); F.Skip(Rep); DD\end{tabular}} & \multicolumn{1}{c|}{opt} \\ \cline{1-13}

\cellcolor{gray!10} & 0\% $(0.00)$ & \cellcolor{green!25}10 & \cellcolor{green!25}15 & \cellcolor{green!25}10 & \multicolumn{1}{c|}{\cellcolor{green!25}1} & \cellcolor{green!25}RND & \cellcolor{green!25}Tourn. & \cellcolor{green!25}Uniform & \cellcolor{green!25}Uniform & \cellcolor{green!25}Last & \multicolumn{1}{|c|}{\cellcolor{green!25}-} & \multicolumn{1}{c|}{\cellcolor{yellow!25}\cite{gaina2017rhanalysis}} \\ \cline{2-13} 
\multirow{-2}{*}{\cellcolor{gray!10}2} & 0\% $(0.00)$ & 1 & 20 & 15 & \multicolumn{1}{c|}{\cellcolor{green!25}1} & MCTS & Rank & \cellcolor{green!25}Uniform & 2-bit & \cellcolor{green!25}Last & \multicolumn{1}{|c|}{\begin{tabular}[c]{@{}c@{}}S.Buff(0.9); MC(0.5,1); F.Skip(RND)\end{tabular}} & \multicolumn{1}{c|}{opt} \\ \cline{1-13}

\cellcolor{gray!10} & 100\% $(0.00)$ & \cellcolor{green!25}10 & \cellcolor{green!25}15 & \cellcolor{green!25}10 & \multicolumn{1}{c|}{\cellcolor{green!25}1} & \cellcolor{green!25}RND & \cellcolor{green!25}Tourn. & \cellcolor{green!25}Uniform & \cellcolor{green!25}Uniform & \cellcolor{green!25}Last & \multicolumn{1}{|c|}{\cellcolor{green!25}-} & \multicolumn{1}{c|}{\cellcolor{yellow!25}\cite{gaina2017rhanalysis}} \\ \cline{2-13} 
\multirow{-2}{*}{\cellcolor{gray!10}3} & 99\% $(0.99)$ & \cellcolor{green!25}10 & \cellcolor{green!25}15 & 20 & \multicolumn{1}{c|}{\cellcolor{green!25}1} & 1SLA & - & - & \cellcolor{green!25}Uniform & Disc. & \multicolumn{1}{|c|}{S.Buff(0.9); MC(1.0,5); DD} & \multicolumn{1}{c|}{opt} \\ \cline{1-13}

\cellcolor{gray!10} & 10\% $(3.00)$ & \cellcolor{green!25}10 & \cellcolor{green!25}15 & \cellcolor{green!25}10 & \multicolumn{1}{c|}{\cellcolor{green!25}1} & \cellcolor{green!25}RND & \cellcolor{green!25}Tourn. & \cellcolor{green!25}Uniform & \cellcolor{green!25}Uniform & \cellcolor{green!25}Last & \multicolumn{1}{|c|}{\cellcolor{green!25}-} & \multicolumn{1}{c|}{\cellcolor{yellow!25}\cite{gaina2017rhanalysis}} \\ \cline{2-13} 
\multirow{-2}{*}{\cellcolor{gray!10}4} & 10\% $(3.00)$ & 20 & 10 & 15 & \multicolumn{1}{c|}{0} & 1SLA & - & - & 2-bit & Disc. & \multicolumn{1}{|c|}{S.Buff(0.9); DD} & \multicolumn{1}{c|}{opt} \\ \cline{1-13}

\cellcolor{gray!10} & \textbf{13.13\% $(3.39)$} & \cellcolor{green!25}10 & \cellcolor{green!25}15 & \cellcolor{green!25}10 & \multicolumn{1}{c|}{\cellcolor{green!25}1} & \cellcolor{green!25}RND & \cellcolor{green!25}Tourn. & \cellcolor{green!25}Uniform & \cellcolor{green!25}Uniform & \cellcolor{green!25}Last & \multicolumn{1}{|c|}{\cellcolor{green!25}-} & \multicolumn{1}{c|}{\cellcolor{yellow!25}\cite{gaina2017rhanalysis}} \\ \cline{2-13} 
\multirow{-2}{*}{\cellcolor{gray!10}5} & 3\% $(1.70)$ & \cellcolor{green!25}10 & 20 & 20 & \multicolumn{1}{c|}{\cellcolor{green!25}1} & \cellcolor{green!25}RND & \cellcolor{green!25}Tourn. & 2-point & - & Average & \multicolumn{1}{|c|}{MC(0.5,1)} & \multicolumn{1}{c|}{opt} \\ \cline{1-13}

\cellcolor{gray!10} & 11.00\% $(3.129)$ & 1 & 10 & \cellcolor{green!25}10 & \multicolumn{1}{c|}{\cellcolor{green!25}1} & \cellcolor{green!25}RND & \cellcolor{green!25}Tourn. & \cellcolor{green!25}Uniform & \cellcolor{green!25}Uniform & \cellcolor{green!25}Last & \multicolumn{1}{|c|}{MC(0.5,10)} & \multicolumn{1}{c|}{\cellcolor{yellow!25}\cite{gaina2017rhhybrids}} \\ \cline{2-13} 
\multirow{-2}{*}{\cellcolor{gray!10}6} & \textbf{41\% $(4.918)$} & 15 & \cellcolor{green!25}15 & 1 & \multicolumn{1}{c|}{\cellcolor{green!25}1} & MCTS & \cellcolor{green!25}Tourn. & \cellcolor{green!25}Uniform & Diversity & Disc. & \multicolumn{1}{|c|}{\begin{tabular}[c]{@{}c@{}}S.Buff(0.99); MC(1.0,5);\\ F.Skip(Rep);  Fit.Div(0.5)\end{tabular}} & \multicolumn{1}{c|}{opt} \\ \cline{1-13}

\cellcolor{gray!10} & \textbf{46\% $(4.984)$} & \cellcolor{green!25}10 & \cellcolor{green!25}15 & \cellcolor{green!25}10 & \multicolumn{1}{c|}{\cellcolor{green!25}1} & \cellcolor{green!25}RND & \cellcolor{green!25}Tourn. & \cellcolor{green!25}Uniform & \cellcolor{green!25}Uniform & \cellcolor{green!25}Last & \multicolumn{1}{|c|}{MC(0.5,1)} & \multicolumn{1}{c|}{\cellcolor{yellow!25}\cite{gaina2017rhhybrids}} \\ \cline{2-13} 
\multirow{-2}{*}{\cellcolor{gray!10}7} & 32\% $(4.665)$ & 20 & 10 & \cellcolor{green!25}10 & \multicolumn{1}{c|}{0} & \cellcolor{green!25}RND & - & - & Diversity & Disc. & \multicolumn{1}{|c|}{S.Buff(0.9); Fit.Div(0.5)} & \multicolumn{1}{c|}{opt} \\ \cline{1-13}

\cellcolor{gray!10} & \textbf{12\% $(3.25)$} & \cellcolor{green!25}10 & \cellcolor{green!25}15 & \cellcolor{green!25}10 & \multicolumn{1}{c|}{\cellcolor{green!25}1} & \cellcolor{green!25}RND & \cellcolor{green!25}Tourn. & \cellcolor{green!25}Uniform & \cellcolor{green!25}Uniform \cellcolor{gray!10} & \cellcolor{green!25}Last & \multicolumn{1}{|c|}{S.Buff(0.9); MC(0.5,10)} & \multicolumn{1}{c|}{\cellcolor{yellow!25}\cite{gaina2017rhhybrids}} \\ \cline{2-13} 
\multirow{-2}{*}{\cellcolor{gray!10}8} & 11\% $(3.129)$ & \cellcolor{green!25}10 & 20 & 15 & \multicolumn{1}{c|}{0} & 1SLA & Rank & 2-point & 2-bit & Max & \multicolumn{1}{|c|}{S.Buff(0.99); MC(2.0,5)} & \multicolumn{1}{c|}{opt} \\ \cline{1-13}

\cellcolor{gray!10} & \textbf{20\% $(4.00)$} & \cellcolor{green!25}10 & \cellcolor{green!25}15 & \cellcolor{green!25}10 & \multicolumn{1}{c|}{\cellcolor{green!25}1} & \cellcolor{green!25}RND & \cellcolor{green!25}Tourn. & \cellcolor{green!25}Uniform & \cellcolor{green!25}Uniform & \cellcolor{green!25}Last & \multicolumn{1}{|c|}{S.Buff(0.9); MC(0.5,10)} & \multicolumn{1}{c|}{\cellcolor{yellow!25}\cite{gaina2017rhhybrids}} \\ \cline{2-13} 
\multirow{-2}{*}{\cellcolor{gray!10}9} & 19\% $(3.923)$ & \cellcolor{green!25}10 & 20 & \cellcolor{green!25}10 & \multicolumn{1}{c|}{\cellcolor{green!25}1} & \cellcolor{green!25}RND & \cellcolor{green!25}Tourn. & 1-point & \cellcolor{green!25}Uniform & Max & \multicolumn{1}{|c|}{S.Buff(0.99); MC(1.0,5)} & \multicolumn{1}{c|}{opt} \\ \cline{1-13}

\cellcolor{gray!10} & 78.33\% $(3.76)$ & \cellcolor{green!25}10 & \cellcolor{green!25}15 & \cellcolor{green!25}10 & \multicolumn{1}{c|}{\cellcolor{green!25}1} & \cellcolor{green!25}RND & \cellcolor{green!25}Tourn. & \cellcolor{green!25}Uniform & \cellcolor{green!25}Uniform & \cellcolor{green!25}Last & \multicolumn{1}{|c|}{\cellcolor{green!25}-} & \multicolumn{1}{c|}{\cellcolor{yellow!25}\cite{gaina2017rhanalysis}} \\ \cline{2-13} 
\multirow{-2}{*}{\cellcolor{gray!10}10} & \textbf{83\% $(3.756)$} & \cellcolor{green!25}10 & 20 & \cellcolor{green!25}10 & \multicolumn{1}{c|}{\cellcolor{green!25}1} & 1SLA & \cellcolor{green!25}Tourn. & 1-point & - & Disc. & \multicolumn{1}{|c|}{ \begin{tabular}[c]{@{}c@{}}S.Buff(0.99); MC(2.0,1); \\ F.Skip(Null); DD\end{tabular}} & \multicolumn{1}{c|}{opt} \\ \cline{1-13}

\cellcolor{gray!10} & 54.55\% $(5.00)$ & \cellcolor{green!25}10 & \cellcolor{green!25}15 & \cellcolor{green!25}10 & \multicolumn{1}{c|}{\cellcolor{green!25}1} & \cellcolor{green!25}RND & \cellcolor{green!25}Tourn. & \cellcolor{green!25}Uniform & \cellcolor{green!25}Uniform & \cellcolor{green!25}Last & \multicolumn{1}{|c|}{\cellcolor{green!25}-} & \multicolumn{1}{c|}{\cellcolor{yellow!25}\cite{gaina2017rhanalysis}} \\ \cline{2-13} 
\multirow{-2}{*}{\cellcolor{gray!10}11} & \textbf{56\% $(4.968)$} & 20 & \cellcolor{green!25}15 & \cellcolor{green!25}10 & \multicolumn{1}{c|}{\cellcolor{green!25}1} & \cellcolor{green!25}RND & \cellcolor{green!25}Tourn. & \cellcolor{green!25}Uniform & \cellcolor{green!25}Uniform & Disc. & \multicolumn{1}{|c|}{\begin{tabular}[c]{@{}c@{}}S.Buff(0.99); MC(2.0,5);\\ F.Skip(RND); DD\end{tabular}} & \multicolumn{1}{c|}{opt} \\ \cline{1-13}

\cellcolor{gray!10} & \textbf{37.5\% $(4.42)$} & \cellcolor{green!25}10 & \cellcolor{green!25}15 & \cellcolor{green!25}10 & \multicolumn{1}{c|}{\cellcolor{green!25}1} & \cellcolor{green!25}RND & \cellcolor{green!25}Tourn. & \cellcolor{green!25}Uniform & \cellcolor{green!25}Uniform & \cellcolor{green!25}Last & \multicolumn{1}{|c|}{\cellcolor{green!25}-} & \multicolumn{1}{c|}{\cellcolor{yellow!25}\cite{gaina2017rhanalysis}} \\ \cline{2-13} 
\multirow{-2}{*}{\cellcolor{gray!10}12} & 25\% $(4.33)$ & \cellcolor{green!25}10 & 20 & 15 & \multicolumn{1}{c|}{0} & \cellcolor{green!25}RND & \cellcolor{green!25}Tourn. & 1-point & Diversity & Max & \multicolumn{1}{|c|}{\begin{tabular}[c]{@{}c@{}}MC(0.5,10);\\ F.Skip(Seq); DD\end{tabular}} & \multicolumn{1}{c|}{opt} \\ \cline{1-13}

\cellcolor{gray!10} & 77.78\% $(4.18)$ & \cellcolor{green!25}10 & \cellcolor{green!25}15 & \cellcolor{green!25}10 & \multicolumn{1}{c|}{\cellcolor{green!25}1} & \cellcolor{green!25}RND & \cellcolor{green!25}Tourn. & \cellcolor{green!25}Uniform & \cellcolor{green!25}Uniform & \cellcolor{green!25}Last & \multicolumn{1}{|c|}{\cellcolor{green!25}-} & \multicolumn{1}{c|}{\cellcolor{yellow!25}\cite{gaina2017rhanalysis}} \\ \cline{2-13} 
\multirow{-2}{*}{\cellcolor{gray!10}13} & \textbf{86\% $(3.47)$} & 15 & 20 & 15 & \multicolumn{1}{c|}{\cellcolor{green!25}1} & \cellcolor{green!25}RND & \cellcolor{green!25}Tourn. & 1-point & Softmax & Max & \multicolumn{1}{|c|}{S.Buff(0.9); MC(2.0,1)} & \multicolumn{1}{c|}{opt} \\ \cline{1-13}

\cellcolor{gray!10} & 98.99\% $(1.00)$ & \cellcolor{green!25}10 & \cellcolor{green!25}15 & \cellcolor{green!25}10 & \multicolumn{1}{c|}{\cellcolor{green!25}1} & \cellcolor{green!25}RND & \cellcolor{green!25}Tourn. & \cellcolor{green!25}Uniform & \cellcolor{green!25}Uniform & \cellcolor{green!25}Last & \multicolumn{1}{|c|}{\cellcolor{green!25}-} & \multicolumn{1}{c|}{\cellcolor{yellow!25}\cite{gaina2017rhanalysis}} \\ \cline{2-13} 
\multirow{-2}{*}{\cellcolor{gray!10}14} & \textbf{100\% $(0.00)$} & 15 & 20 & 1 & \multicolumn{1}{c|}{\cellcolor{green!25}1} & 1SLA & Rank & 1-point & Diversity & Max & \multicolumn{1}{|c|}{ \begin{tabular}[c]{@{}c@{}}S.Buff(1.0); MC(2.0,1); F.Skip(RND); DD\end{tabular}} & \multicolumn{1}{c|}{opt} \\ \cline{1-13}

\cellcolor{gray!10} & 65\% $(4.77)$ & 1 & 5 & 1 & \multicolumn{1}{c|}{\cellcolor{green!25}1} & \cellcolor{green!25}RND & \cellcolor{green!25}Tourn. & \cellcolor{green!25}Uniform & \cellcolor{green!25}Uniform & \cellcolor{green!25}Last & \multicolumn{1}{|c|}{S.Buffer(0.9) MC(0.5,10)} & \multicolumn{1}{c|}{\cellcolor{yellow!25}\cite{gaina2017rhhybrids}} \\ \cline{2-13} 
\multirow{-2}{*}{\cellcolor{gray!10}15} & \textbf{84\% $(3.66)$} & 20 & 20 & 1 & \multicolumn{1}{c|}{\cellcolor{green!25}1} & \cellcolor{green!25}RND & Rank & \cellcolor{green!25}Uniform & - & Average & \multicolumn{1}{|c|}{S.Buff(0.9); MC(2.0,1)} & \multicolumn{1}{c|}{opt} \\ \cline{1-13}

\cellcolor{gray!10} & \textbf{100\% $(0.00)$} & \cellcolor{green!25}10 & \cellcolor{green!25}15 & \cellcolor{green!25}10 & \multicolumn{1}{c|}{\cellcolor{green!25}1} & \cellcolor{green!25}RND & \cellcolor{green!25}Tourn. & \cellcolor{green!25}Uniform & \cellcolor{green!25}Uniform & \cellcolor{green!25}Last & \multicolumn{1}{|c|}{\cellcolor{green!25}-} & \multicolumn{1}{c|}{\cellcolor{yellow!25}\cite{gaina2017rhanalysis}} \\ \cline{2-13} 
\multirow{-2}{*}{\cellcolor{gray!10}16} & 99\% $(0.99)$ & 15 & 20 & 1 & \multicolumn{1}{c|}{0} & \cellcolor{green!25}RND & - & - & 2-bit & \cellcolor{green!25}Last & \multicolumn{1}{|c|}{\begin{tabular}[c]{@{}c@{}}S.Buff(0.99);  MC(0.5,10); F.Skip(RND); DD\end{tabular}} & \multicolumn{1}{c|}{opt} \\ \cline{1-13}

\cellcolor{gray!10} & 100\% $(0.00)$ & \cellcolor{green!25}10 & \cellcolor{green!25}15 & \cellcolor{green!25}10 & \multicolumn{1}{c|}{\cellcolor{green!25}1} & \cellcolor{green!25}RND & \cellcolor{green!25}Tourn. & \cellcolor{green!25}Uniform & \cellcolor{green!25}Uniform & \cellcolor{green!25}Last & \multicolumn{1}{|c|}{\cellcolor{green!25}-} & \multicolumn{1}{c|}{\cellcolor{yellow!25}\cite{gaina2017rhanalysis}} \\ \cline{2-13} 
\multirow{-2}{*}{\cellcolor{gray!10}17} & 100\% $(0.00)$ & 15 & 20 & 20 & \multicolumn{1}{c|}{0} & \cellcolor{green!25}RND & - & - & 2-bit & Disc. & \multicolumn{1}{|c|}{S.Buff(0.9); DD} & \multicolumn{1}{c|}{opt} \\ \cline{1-13}

\cellcolor{gray!10} & \textbf{96\% $(1.92)$} & \cellcolor{green!25}10 & \cellcolor{green!25}15 & \cellcolor{green!25}10 & \multicolumn{1}{c|}{\cellcolor{green!25}1} & \cellcolor{green!25}RND & \cellcolor{green!25}Tourn. & \cellcolor{green!25}Uniform & \cellcolor{green!25}Uniform & \cellcolor{green!25}Last & \multicolumn{1}{|c|}{\cellcolor{green!25}-} & \multicolumn{1}{c|}{\cellcolor{yellow!25}\cite{gaina2017rhanalysis}} \\ \cline{2-13} 
\multirow{-2}{*}{\cellcolor{gray!10}18} & 90\% $(3.0)$ & 15 & 5 & 20 & \multicolumn{1}{c|}{0} & \cellcolor{green!25}RND & Roulette & 2-point & - & Disc. & \multicolumn{1}{|c|}{S.Buff(0.99); MC(2.0,5)} & \multicolumn{1}{c|}{opt} \\ \cline{1-13}

\cellcolor{gray!10} & 100\% $(0.00)$ & \cellcolor{green!25}10 & \cellcolor{green!25}15 & \cellcolor{green!25}10 & \multicolumn{1}{c|}{\cellcolor{green!25}1} & \cellcolor{green!25}RND & \cellcolor{green!25}Tourn. & \cellcolor{green!25}Uniform & \cellcolor{green!25}Uniform & \cellcolor{green!25}Last & \multicolumn{1}{|c|}{\cellcolor{green!25}-} & \multicolumn{1}{c|}{\cellcolor{yellow!25}\cite{gaina2017rhanalysis}} \\ \cline{2-13} 
\multirow{-2}{*}{\cellcolor{gray!10}19} & 100\% $(0.00)$ & \cellcolor{green!25}10 & 5 & 1 & \multicolumn{1}{c|}{\cellcolor{green!25}1} & 1SLA & Roulette & 1-point & 2-bit & Disc. & \multicolumn{1}{|c|}{MC(2.0,5); DD} & \multicolumn{1}{c|}{opt} \\ \hline
 
\end{tabular}
\label{tab:perf}
\end{table*}

\ifCLASSOPTIONcaptionsoff
  \newpage
\fi

\bibliographystyle{IEEEtran} 
\bibliography{IEEEabrv,bibs/EAhybrids_refs,bibs/EAnovelty_refs,bibs/EAseeding_refs,bibs/MO_refs,bibs/agents-EA_refs,bibs/agents-MCTS_refs,bibs/agents-RL_refs,bibs/agents-general_refs,bibs/macro_refs,bibs/competition_refs,bibs/gamedesign_refs,bibs/agents-analysis_refs,bibs/agents-heuristic_refs,bibs/other_refs,bibs/ntbea_refs}

\end{document}